\documentclass[journal]{IEEEtran}

\usepackage{makecell}
\usepackage{boldline}
\usepackage{caption}
\usepackage{booktabs} 

\usepackage{url}
\usepackage{xcolor}
\definecolor{newcolor}{rgb}{.8,.349,.1}

\usepackage{moreverb,textcomp,graphicx,multirow,setspace}
\usepackage{amssymb,amsmath,mathptmx}
\usepackage{subfigure,psfrag,epsfig}
\usepackage{color,helvet,courier,type1cm,makeidx,multicol}
\usepackage[bottom]{footmisc}
\usepackage{bm,amsfonts,rotating,lineno,url,times}
\usepackage[algoruled,vlined,linesnumbered]{algorithm2e}
\usepackage{color} 
\newcommand\newnotecommand[3]{%
  \newcommand#1[1]{{\color{#3}\footnote{{\color{#3}#2:} ##1}}}}
\newnotecommand\joni{Joni}{red}
\newnotecommand\cory{Cory}{blue}

\usepackage{paralist} 
\newenvironment{packed_itemize}{
\begin{itemize}
  \setlength{\itemsep}{1pt}
  \setlength{\parskip}{0pt}
  \setlength{\parsep}{0pt}
}{\end{itemize}}
\def\onedot{. }
\def\eg{\emph{e.g}\onedot} 
\def\ie{\emph{i.e}\onedot}

\def\etal{\emph{et al}\onedot}

\newcommand{\mat}[1]{\boldsymbol{#1}}

\begin{document}

%
%
%

\title{Convolutional Low-Resolution Fine-Grained Classification}

\author{Dingding Cai, Ke Chen, Yanlin Qian, Joni-Kristian K\"am\"ar\"ainen}

\maketitle

\begin{abstract}
Successful fine-grained image classification methods learn subtle details between visually similar (sub-)classes, but the problem becomes significantly more challenging if the details are missing due to low resolution. Encouraged by the recent success of Convolutional Neural Network (CNN) architectures in image classification, we propose a novel resolution-aware deep model which combines convolutional image super-resolution and convolutional fine-grained classification into a single model in an end-to-end manner. Extensive experiments on multiple benchmarks demonstrate that the proposed model consistently performs better than conventional convolutional networks on classifying fine-grained object classes in low-resolution images. 
\end{abstract}

\begin{IEEEkeywords}
 Fine-Grained Image classification, Super Resolution Convoluational Neural Networks, Deep Learning 
\end{IEEEkeywords}

\IEEEpeerreviewmaketitle

The problem of image classification is to categorise images according
to their semantic content (\eg person, plane). Fine-grained image
classification further divides classes to their ``sub-categories''
such as the models of cars~\cite{krause20133d}, the species of birds
\cite{wah2011caltech},  the categories of flowers \cite{nilsback2008automated} and the breeds of dogs \cite{khosla2011novel}.
Fine-grained categorisation is a difficult task due to small
inter-class variance between visually similar sub-classes. The problem
becomes even more challenging when available images are
low-resolution (LR) images where many details are missing as
compared to their high-resolution (HR) counterparts.

Since the rise of Convolutional Neural Network (CNN) architectures
in image classification~\cite{krizhevsky2012imagenet}, the accuracy
of fine-grained image classification has dramatically improved
and many CNN-based extensions have been proposed
\cite{zhang2014part,lin2015bilinear,krause2015fine,Chen2016tbd,akata2015evaluation,BraHorBel:2014}.
However, these works assume sufficiently good image quality and
high resolution, (\eg typically $227\times 227$
for AlexNet~\cite{krizhevsky2012imagenet}) while with low resolution
images the CNN performance quickly
collapses~\cite{chevalier2015lr,Liu_ICPRWS-2016}.
The challenge raises from the problem of how to recover necessary
texture details from low-resolution images. Our solution is to
adopt image super-resolution (SR) techniques
\cite{zeyde2010single,yang2010image,chang2004super,glasner2009super,dong2016image}
to enrich imagery details. In particular, inspired by the recent
work on CNN-based image SR by Deng \etal \cite{dai2016image} we
propose a unique end-to-end deep learning framework that combines
CNN super-resolution and CNN fine-grained classification -- a resolution-aware Convolutional Neural Network (RACNN) for
fine-grained object categorisation in low-resolution images. 
To our best knowledge, our work is the first end-to-end learning model for low-resolution fine-grained object classification.

\begin{figure}[t]
\centering
\includegraphics[width=0.98\linewidth]{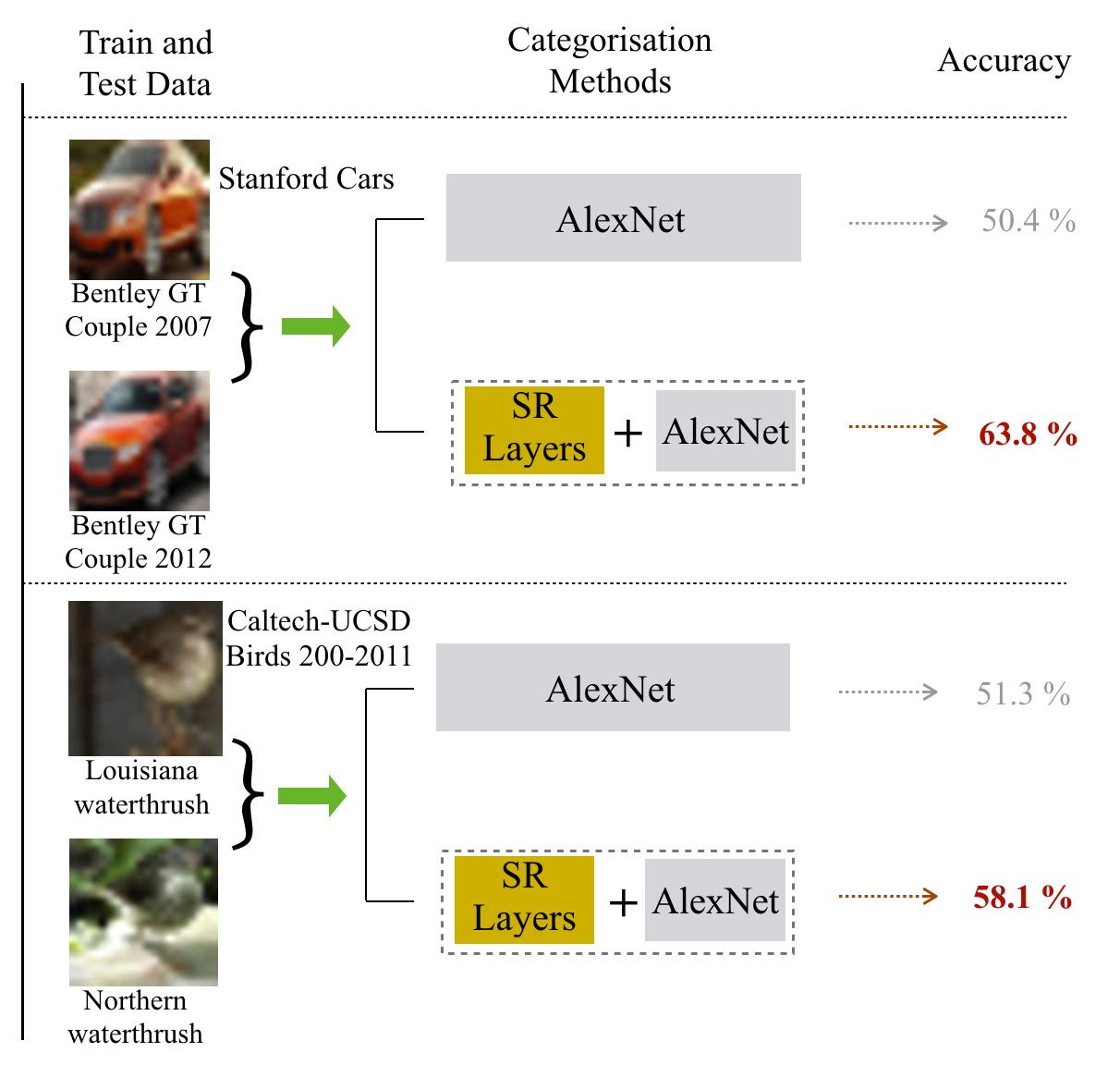}
\caption{Owing to the introduction of the convolutional super-resolution
  (SR) layers, the proposed deep convolutional model (the bottom pipelines) achieves
  superior performance for low resolution images.} 
\label{fig:intro}
\end{figure}

Our main principle is simple: the higher
image resolution, the easier for classification. Our research
questions are: Can computational super-resolution recover some of
the important details required for fine-grained image classification
and can such SR layers be added to an end-to-end deep classification
architecture?  
To this end, our RACNN integrates
deep residual learning for image super-resolution
\cite{kim2016accurate} into typical convolutional classification networks (\eg AlexNet
\cite{krizhevsky2012imagenet} or VGG-Net \cite{Simonyan15}). On one hand, the proposed RACNN has deeper
network architecture (\ie more network parameters) than the
straightforward solution of conventional CNN on
up-scaled images (\eg bicubic interpolation \cite{keys1981cubic}).
Our RACNN learns to refine and provide more texture details
for low-resolution images to boost fine-grained classification performance. 
We conduct experiments on three benchmarks, Stanford
Cars~\cite{krause20133d}, Caltech-UCSD Birds-200-2011~\cite{wah2011caltech} and
 Oxford 102 Flower Dataset~\cite{nilsback2008automated}.
Our results answer the aforementioned questions: super-resolution improves
fine-grained classification and SR-based fine-grained classification
can be designed into a supervised end-to-end learning framework, as depicted in
Figure~\ref{fig:intro} illustrating the difference
between RACNN and conventional CNN.

\section{Related Work}

\vspace{\medskipamount} \noindent {\bf Fine-Grained Image Categorisation -- }
Recent algorithms for discriminating fine-grained classes (such as
animal species or plants
\cite{wah2011caltech,khosla2011novel,angelova2013efficient} and man-made
objects \cite{krause20133d,stark2011fine,maji2013fine}) can be
divided into two main groups. 
The first group of methods utilises discriminative visual
cues from local parts obtained by detection
\cite{zhang2014part,zhang2013deformable} or segmentation
\cite{krause2015fine, chai2013symbiotic,gavves2015local}. 
The second group of methods focuses on discovering inter-class
label dependency via pre-defined hierarchical structure of labels
\cite{akata2015evaluation,shotton2008semantic, 
hwang2012semantic,mittal2012taxonomic,deng2014large} or
manually-annotated visual attributes
\cite{zhang2014panda,fu2014transductive}.  
Significant performance improvement is achieved by convolutional
neural networks (CNNs), but this 
requires a massive amount of high
quality training images. Fine-grained classification from low-resolution images
is yet challenging and unexplored.
The method proposed by Peng \etal \cite{peng2016fine} transforms detailed texture
information in HR images to LR via fine-tuning to boost the accuracy of
recognizing fine-grained objects in LR images. 
However, in \cite{peng2016fine}, their strong assumption requiring HR images available for
training limits its generalisation
ability.  In addtion, the same assumption also occurs in Wang's work \cite{wang2016studying}.
Chevalier \etal \cite{chevalier2015lr} design a CNN-based
fine-grained object classifier with respect to varying image
resolutions, which adopts ordinary convolutional and fully-connected
layers but misses considering super-resolution specific layers in convolutional classification networks.    
On contrary, owing to the introduction of SR-specific layers in RACNN,
our method can consistently gain notable performance improvement over
conventional CNN for image classification on fine-grained
classification datasets. 

\vspace{\medskipamount} \noindent {\bf Convolutional Super-Resolution Layers -- }
Yang \etal \cite{yang2014single} grouped existing SR algorithms into
four groups: prediction models \cite{irani1991improving}, edge-based
methods \cite{fattal2007image}, image statistical methods
\cite{huang1999statistics} and example-based methods
\cite{zeyde2010single,glasner2009super,dong2016image,huang2015single, 
yang2013fast,freedman2011image,dai2015jointly,schulter2015fast}. 
Recently, Convolutional Neural Networks have been adopted for image
super-resolution achieving state-of-the-art performance.   
The first attempt using convolutional neural networks for image
super-resolution was proposed by Dong
\etal \cite{dong2016image}. Their
method learns a deep mapping between low- and high- resolution patches
and has inspired a number of follow-ups
\cite{kim2016accurate,kim2016deeply,dong2016accelerating}. 
In \cite{dong2016accelerating}, an additional deconvolution layer is
added based on SRCNN \cite{dong2016image} to avoid general up-scaling
of input patches for accelerating CNN training and testing. 
Kim \etal \cite{kim2016deeply} adopt a deep recursive layer to avoid
adding weighting layers, which does not need to pay any price of
increasing network parameters. 
In \cite{kim2016accurate}, a convolutional deep network is proposed to
learn the mapping between LR image and its residue between LR and HR
image to speed up CNN training for very deep network.   
Convolutional layers designed for image super-resolution
(namely SR-specific convolutional layers) have been verified their
effectiveness to improve the quality of images.  
In this work, we incorporate the state-of-the-art residual CNN
layers for image super-resolution \cite{kim2016accurate} into a
convolutional categorisation network for classifying
fine-grained objects (\ie AlexNet \cite{krizhevsky2012imagenet}, 
VGG-Net \cite{Simonyan15} and  GoogLeNet \cite{szegedy2015going}). 
In the experiments, SR-specific convolutional layers are verified to improve classification performance.



\vspace{\medskipamount} \noindent {\bf Contributions -- }
Our contributions are two-fold: 
\begin{packed_itemize}
\item Our work is the first attempt to utilise super-resolution
  specific convolutional layers to improve convolutional fine-grained
  image classification. 
\item We experimentally verify that the proposed RACNN achieves superior
  performance on low-resolution images which
  make ordinary CNN performance collapse.  
\end{packed_itemize}


\begin{figure*}[t]
\centering
\includegraphics[width=0.98\linewidth]{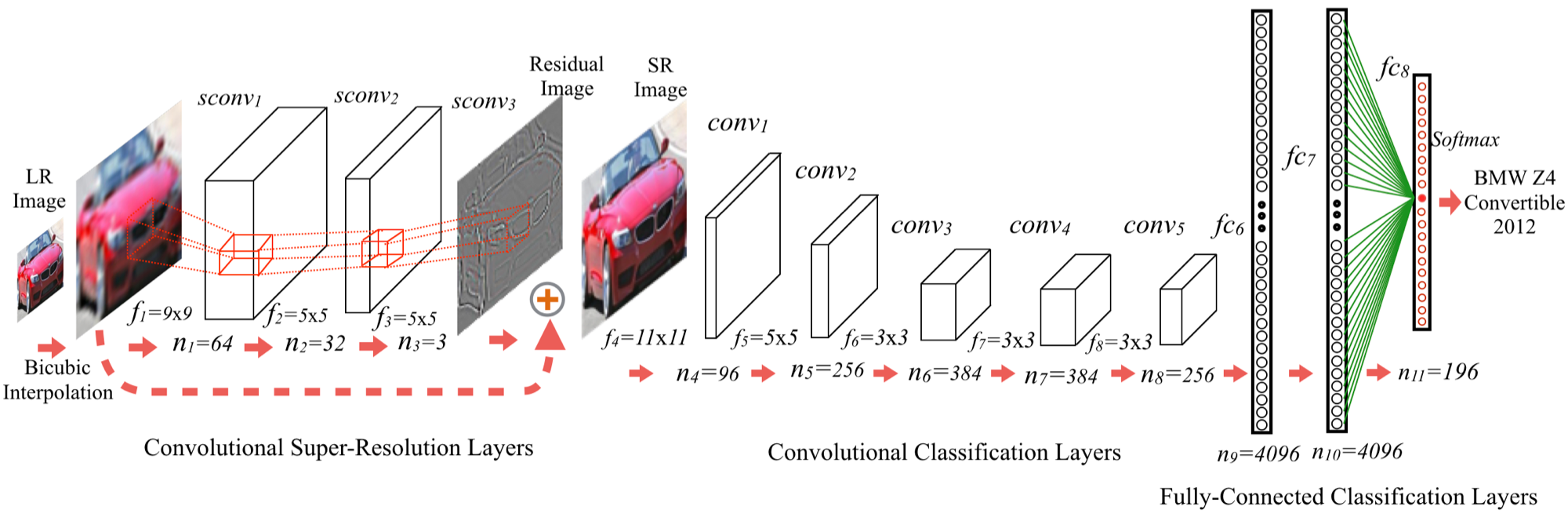}
\caption{Pipeline of the proposed resolution-aware convolutional
  neural network (RACNN) for fine-grained recognition with
  low-resolution images. Convolutional classification layers from
  AlexNet are adopted for illustrative purpose, which can be readily
  replaced by those from other CNNs such as VGG-Net or GoogLeNet}.
\label{fig:pipline}
\end{figure*}

\section{Resolution-Aware Convolutional Neural Networks}

Given a set of $N$ training images and corresponding class labels
$\{\mat{X}_i,y_i\}, i=1,2,\cdots,N$, the goal of a conventional
CNN model is to learn a mapping function $y=f(\mat{X})$. 
The typical cross entropy (ce) loss $L_\text{ce}(\cdot)$ on softmax classifer is adopted to
measure the performance between class estimates $\hat{y} = f(\mat{X})$
and ground truth class labels $y$ :  
\begin{equation}
L_\text{ce}(\hat{y},y)= -\sum_{j=1}^ly_j\log(\hat{y}_j),
\end{equation}
where $j$  refers to the
index of element in vectors, and $l$ denotes the dimension of softmax
layer (\ie the number of classes). 
In this sense, CNN solves the following minimisation problem with
gradient descent back propagation: 
\begin{equation}
\min ~~\sum_{i=1}^N L_\text{ce}(f(\mat{X}_i),y_i).
\label{eqn.cross_entropy}
\end{equation}
For fine-grained categorisation in low-quality images, we propose a
novel resolution-aware convolutional neural network, which is
illustrated in Fig.~\ref{fig:pipline}. 
In general, our RACNN consists of two parts: convolutional
super-resolution layers (see Sec. \ref{subsec.srLayers}) and
convolutional categorisation layers (see
Sec. \ref{subsec.cateLayers}).  
In Sec. \ref{subsec.training}, we describe an end-to-end training scheme for the
proposed RACNN.
   

\subsection{Convolutional Super-Resolution Layers}\label{subsec.srLayers}

In this section, we present convolutional super-resolution
specific layers for the resolution-aware CNN, the goal of which is to
recover texture details of low-resolution images to feed into the
following convolutional categorisation layers.  

We first investigate the conventional CNN for the super-resolution
task. Given $K$ training pairs of low-resolution and high-resolution
images $\{\mat{X}^\text{LR},\mat{X}^\text{HR}\}^{i}, i =1,2,\cdots,K$,
a direct CNN-based mapping function $g(\mat{X}^\text{LR})$ from
$\mat{X}^\text{LR}$ (input observation) to $\mat{X}^\text{HR}$ (output
target)  \cite{dong2016image,dong2016accelerating} is learned by
minimising the mean square (ms) loss 
\begin{equation}
L_\text{ms}(\mat{X}^\text{LR},\mat{X}^\text{HR}) = \frac{1}{2} \sum_{i=1}^K \|\mat{X}^\text{HR} - g(\mat{X}^\text{LR})\|^2.
\end{equation}

Inspired by recent the state-of-the-art residual convolutional network
\cite{kim2016accurate} to achieve high efficacy, 
we design convolutional super-resolution layers as shown on the left hand
side of Fig.~\ref{fig:pipline}. 
Similar to \cite{kim2016accurate}, our convolutional super-resolution
layers learn a mapping function from LR images
$\mat{X}^\text{LR}$ to residual images $\mat{X}^\text{HR} -
\mat{X}^\text{LR}$.  
Object function of the proposed convolutional super-resolution layers
is as the following: 
\begin{equation}
\min ~~\frac{1}{2} \sum_{i=1}^K \|\mat{X}^\text{HR} - \mat{X}^\text{LR} - g(\mat{X}^\text{LR})\|^2.\label{eqn.rsr}
\end{equation}
The better performance of residual learning yields from the fact that,
since the input (LR) and output images (HR) are largely similar, it is more meaningful to learn their residue where
similarities are removed. It is obvious that detailed
imagery information in the form of residual images is easier for CNNs
to learn than direct LR-HR CNN models~\cite{dong2016image,dong2016accelerating}.  

We utilise three typical stacked convolutional-ReLU layers with
zero-padding filters as convolutional SR layers in RACNN. Following \cite{dong2016image},  
the empirical basic setting of the layers is $f_{1} = 9\times 9$,
$n_{1} = 64$, $f_{2} = 5\times 5$, $n_{2} = 32$, $f_{3} = 5\times 5$
and $n_{3} = 3$, which are also illustrated in the left hand side of
Fig.~\ref{fig:pipline}, where $f_{m}$ and $n_{m}$ donate the size and
number of the filters of the $m$th layer respectively.  
The output of the last convolutional SR layer is summed with the
low-resolution input image $\mat{X}^\text{LR}$ to construct the
full super-resolution image
fed into the remaining convolutional and fully-connected classification layers of RACNN. 

\subsection{Categorisation Layers}\label{subsec.cateLayers}

The second part in our RACNN is convolutional and fully-connected
classification layers with high quality images after super-resolution
layers.  
A number of CNN
frameworks~\cite{krizhevsky2012imagenet,Simonyan15,he2016deep,szegedy2015going}
have been proposed for image categorisation, and in this paper we
consider  three popular convolutional neural networks: AlexNet
\cite{krizhevsky2012imagenet}, VGG-Net \cite{Simonyan15}  and GoogLeNet \cite{szegedy2015going}. 
 All CNNs typically consist of a number of Convolutional-ReLU-Pool
stacks followed by several fully-connected layers. 
On the right-hand-side of Fig. \ref{fig:pipline}, the typical AlexNet
\cite{krizhevsky2012imagenet} is visualised and employed as
convolutional categorisation layers in RACNN. 
AlexNet \cite{krizhevsky2012imagenet}, the baseline CNN for large-scale
image classification over ImageNet \cite{deng2009imagenet}, consists
of 5 convolutional layers (\ie \textit{conv1}, \textit{conv2},
\textit{conv3}, \textit{conv4}, and \textit{conv5}) and 3
fully-connected layers (\ie \textit{fc6}, \textit{fc7}, and
\textit{fc8}).  
VGGNet \cite{Simonyan15} is made deeper (\ie from 8 layers of
Alexnet to 16-19 layers) and more advanced over AlexNet 
by using very small (\eg $3\times 3$) convolution filters.
In our paper, we choose the VGG-Net-16 with 16 layers for our
experiments (denoted as VGG-Net in the rest of the paper).   GoogLeNet \cite{szegedy2015going} comprises 22 layers but has much less number of parameters than AlexNet and VGG-Net owing to the  smaller amount of weights of fully-connected layers.
GoogLeNet generally generates three outputs at various depths for each input, but for simplicity only the last output (\ie the deepest output) is considered in our experiments.
In our experiments,   all three networks are pre-trained on the
Imagenet data and fine-tuned with $\{\mat{X},y\}$ from fine-grained
data as the baseline. 
For fair comparison, we fine-tune the identical pre-trained CNN models
as our convolutional categorisation layers with $\{\mat{X},y\}$ by
replacing the dimension of final fully-connected layer with the size
of object classes $l$.



\subsection{Network Training}\label{subsec.training}
\begin{figure}[t]
\centering

\includegraphics[width=0.95\linewidth]{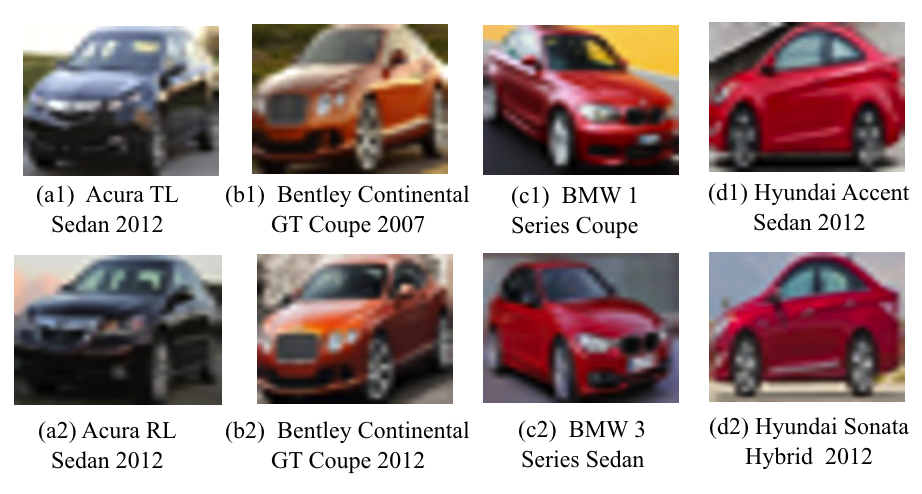}
\includegraphics[width=0.98\linewidth]{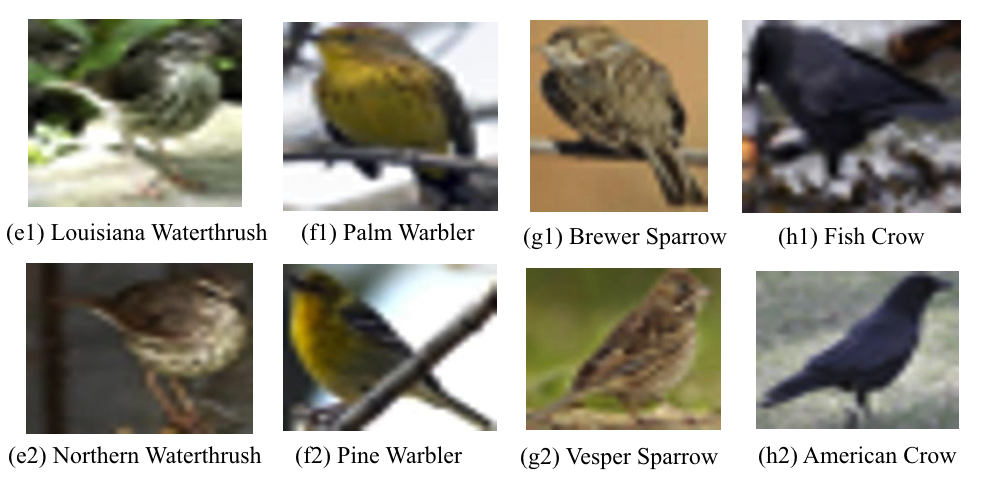}
\caption{Low-resolution image samples after removing background from
  the Stanford Cars and UCSD-Caltech Birds 200-2011 benchmarks.} 
\label{fig:cars}
\end{figure}

The key difference between the proposed resolution-aware CNN and conventional CNN lies in the introduction of three
convolutional super-resolution layers. 
Evidently, RACNN is deeper than corresponding CNN due to the
three convolutional SR layers, which can store more knowledge,
\ie network parameters.  
Before learning RACNN in an end-to-end fashion, we consider two weight
initialization strategies for convolutional SR layers in RACNN,   
\ie standard Gaussian weights and pre-trained weights on the ImageNet
data. 
For fair comparison, we adopt the identical network structure for both initialisation schemes.

For RACNN with Gaussian initial weights, we train the whole network to minimise cross-entropy loss (\ref{eqn.cross_entropy}) directly.
During training, we set learning rates $1$ and weight decays $0.1$ for the first two SR layers
(\textit{sconv1} and \textit{sconv2}) and both learning rate and
weight decay are set with $0.1$ for the third convolutional SR layer
(\textit{sconv3}), while learning rates and
weight decays are $0.1$ and $0$ for all categorisation layers except the last fully-connected layer which
uses both learning rate and weight decay $1$.  

We consider an alternative initialisation strategy for better initial weights for convolutional SR layers.  
To this end, we pre-train the
three convolutional SR layers by enforcing the minimal of the mean square loss
(\ref{eqn.rsr}) on ILSVRC 2015 ImageNet object detection testing
dataset \cite{ILSVRC15}, which consists of 11,142 high-resolution images. 
Given the pre-trained weights in convolutional SR layers, 
RACNN is end-to-end trained by minimising the loss function (\ref{eqn.cross_entropy}) for categorisation. 
For the goal of direct utilisation of output of convolutional SR
layers,  we train SR layers in RGB color space with all the channels,
instead of only on luminance channel Y in YCbCr color space
\cite{dong2016image}. 
Specifically, we generate LR images from HR images (\eg 227$\times$227 pixels) via firstly
down-sampling HR images to 50$\times$50 pixels and then up-scaling
to the original image size by bicubic interpolation
\cite{keys1981cubic}.  
We then sample image patches using sliding window and thus obtain
thousands of pairs of LR and HR image patches. 
To be consistent with the setting of RACNN using Guassian initial weights, the super-resolution layers are trained with image patches by setting
learning rates being $1$ and weight decays being $0.1$ for the first
two SR layers (\textit{sconv1} and \textit{sconv2}) and both learning
rate and weight decay being $0.1$ for the third SR layer
(\textit{sconv3}).  
Finally, we jointly learn both convolutional SR and classification
layers in an end-to-end learning manner with learning rates $0.1$ and
weight decays $0$ for all classification layers except the last fully-connected layer
with both learning rate and weight decay set to $1$. 
 

\section{Experiments}

\subsection{Datasets and Settings}
We evaluate RACNN on  three commonly-used datasets: the
Stanford Cars~\cite{krause20133d}, the Caltech-UCSD
Birds-200-2011~\cite{wah2011caltech} and  the Oxford 102 Category
Flower~\cite{nilsback2008automated} datasets.  
The first one was released by Krause \etal for fine-grained
categorisation and contains 16,185 images from 196 classes of
cars and each class is typically at the level of Brand, Model and Year.  
By following the standard evaluation protocol \cite{krause20133d},  we
split the data into 8,144 images for training and 8,041 for testing.  
{Caltech-UCSD Birds-200-2011} is another challenging fine-grained
image dataset aimed at subordinate category classification by
providing a comprehensive set of benchmarks and annotation types for
the domain of birds. The dataset contains 11,788 images of 200 bird
species, among which there are 5,994 images for training and 5,794 for
testing~\cite{wah2011caltech}.  Oxford 102 Category Flower Dataset consists of
8,189 images which commonly appear in the United Kingdom. These images belong to 102 categories
and each category contains between 40 to 258 images. 
In the standard evaluation protocol~\cite{nilsback2008automated}, the whole dataset
is divided into 1,020 images for training, 1,020 for validation and 6,149 for testing.
In our experiments the training and validation data are merged together to train the networks.

Images from these datasets are first cropped with provided bounding
boxes to remove the background. 
Cropped images are down-sampled to LR images of the size 
$50\times 50$ pixels and then up-scaled to $227\times 227$ pixels by
bicubic interpolation \cite{keys1981cubic} to fit the conventional CNN, which follows the settings in \cite{peng2016fine}.
Sample LR images from the both benchmarks are illustrated in
Fig. \ref{fig:cars}, which verify our motivation to mitigate the
suffering from low visual discrimination due to low-resolution.  
We compare our RACNN with multiple state-of-the-art methods, the
corresponding CNN model for classification (\ie AlexNet
\cite{krizhevsky2012imagenet}, VGG-Net \cite{Simonyan15}  and GoogLeNet \cite{szegedy2015going}) and
Staged-Training CNN proposed by \cite{peng2016fine}.  
The proposed RACNN is implemented on Caffe \cite{jia2014caffe}.  
We adopt {\em the average per-class accuracy}~\cite{Liu_ICPRWS-2016,
  peng2016fine} for the both datasets (the higher value denotes the better performance).

  In our experiments, we used a Lenovo Y900 desktop with one Intel i7-6700K CPU
  and one Nvidia GTX-980 GPU. The proposed RACNN has deeper structure than the competing
  networks (\ie AlexNet, VGGNet, GoogLeNet) which requires longer training times as indicated in
  Table~\ref{table:time}.

\begin{table}[h]
\caption{ Training times of RACNNs and competing CNNs (seconds / epoch)}.
\label{table:time}
\begin{center}
  \resizebox{0.8\linewidth}{!}{
    \begin{tabular}{lrrr}
\toprule
 Methods & ~~~~Cars \cite{krause20133d}  &~~~~Birds \cite{wah2011caltech}  & ~~~~Flowers \cite{nilsback2008automated} \\
\midrule
 AlexNet           & 11 & 8 &  3 \\ 
RACNN$_{\text{AlexNet}}$  & 111 & 80 &  25 \\
\midrule
VGGNet         & 133  &  82 & 34 \\ 
RACNN$_{\text{VGGNet}}$ & 356 & 215 & 90 \\
\midrule
GoogLeNet           &  20 &  25& 8\\ 
RACNN$_{\text{GoogLeNet}}$ & 136 & 120 & 59\\
\bottomrule
\end{tabular}
}
\end{center}
\end{table}

\subsection{Comparative Evaluation}\label{subsec.inter-evaluation}
\begin{figure}[h]
  \centering
  \includegraphics[width=0.98\linewidth]{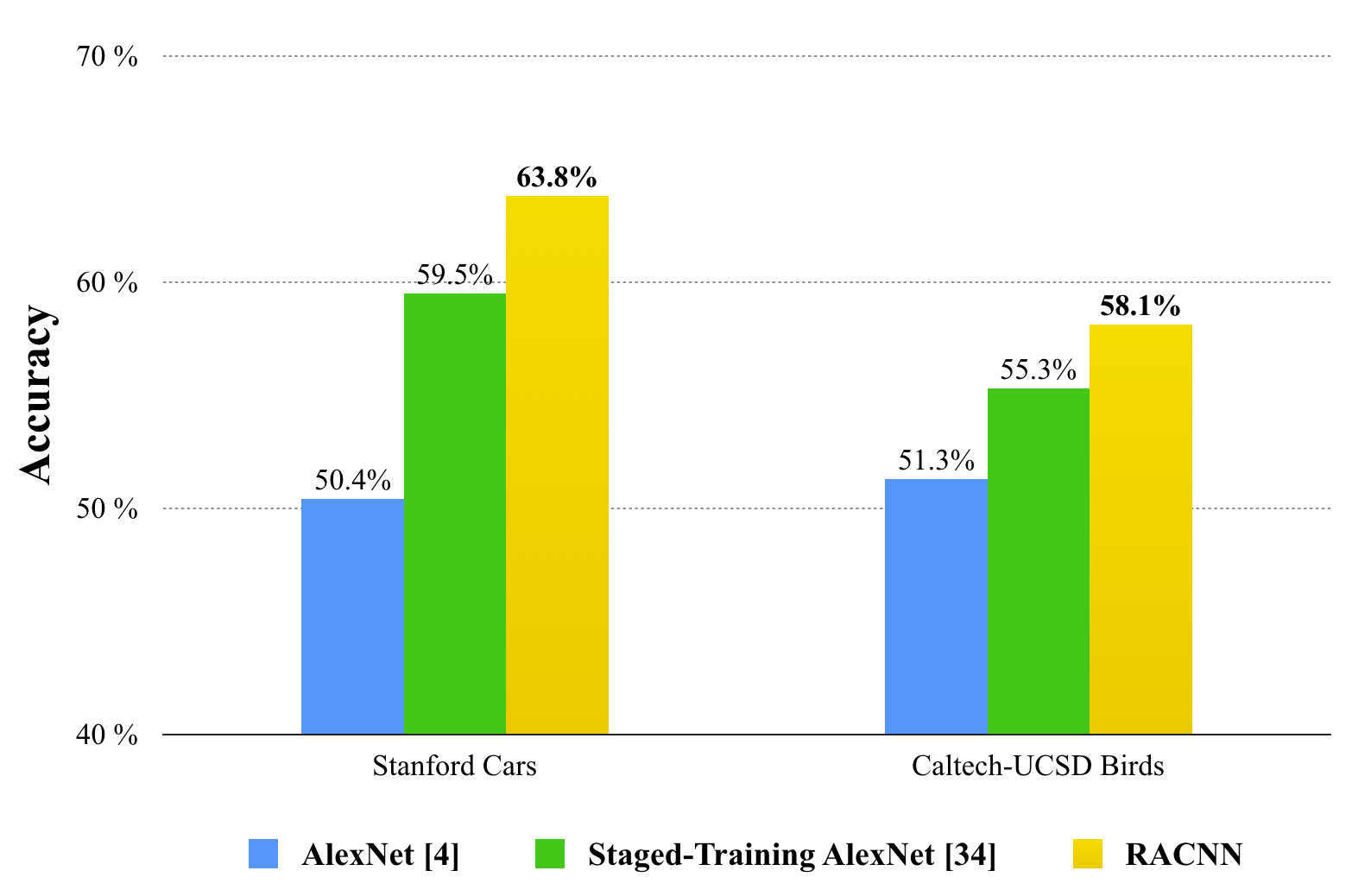}
  \caption{Comparison to two state-of-the-art methods for classification
    (average per-class accuracies).} 
  \label{fig:comparison}
\end{figure}
In  Fig.~\ref{fig:comparison}, we compare our results with AlexNet
\cite{krizhevsky2012imagenet} and Staged-Training AlexNet
\cite{peng2016fine} for fine-grained classification in low-resolution
images.  
It is evident that our RACNN$_\text{AlexNet}$ consistently achieves
the best performance on both benchmarks. 
Precisely, AlexNet achieves 50.4\% and 51.3\% accuracies (collected
from \cite{peng2016fine}) for the Stanford Cars and Caltech-UCSD Birds
datasets, respectively.  
Knowledge transfer between varying resolution images (\ie
Staged-Training AlexNet \cite{peng2016fine}) can improve
classification accuracy, that is  59.5\% for the Stanford Cars and
55.3\% for the Caltech-UCSD Birds. 
However, the staged-training AlexNet \cite{peng2016fine} relies on the
strong assumption that high-resolution images are available for training,
which limits to its usage to other tasks.
Note that our method is more generic and transforms knowledge of super resolution across datasets, which indicates that our method can be readily applied to other low-resolution image classification tasks.   
The proposed RACNN$_\text{AlexNet}$ significantly beats its direct
competitor AlexNet, \ie 63.8\% vs. 50.4\% on the Stanford Cars dataset
and 58.1\% vs. 51.3\% on the Caltech-UCSD Birds dataset. 
With the same settings and training samples, the performance gap can
only be explained by the novel network structure of RACNN. 




\subsection{Evaluation of Convolutional SR Layers}\label{subsec.intra-evaluation}
\begin{table}[t]

  \caption{Evaluation on effect of convolutional SR layers to recover high resolution
    details. We fix all convolutional and fully-connected layers
    except the last fully-connected layer (\ie extracted features correspond to those
    with high resolution images). g-RACNN and p-RACNN denote
    the proposed RACNN with weights initialized with Gaussian (g-) and pre-trained
    weights (p-) for the convolutional SR layers.}
 \label{table:sr-layer}
\begin{center}
{\small \addtolength{\tabcolsep}{-0pt}
\begin{tabular}{lrrrr}
\toprule
Methods & Cars \cite{krause20133d}  & Birds \cite{wah2011caltech}  & Flowers \cite{nilsback2008automated} \\
\midrule
AlexNet \cite{krizhevsky2012imagenet} & 43.75\%& 44.99\%& 70.03\% \\  
g-RACNN$_\text{AlexNet}$  & 45.77\%& 47.17\%&  ~71.91\%\\ 
p-RACNN$_\text{AlexNet}$  & \textbf{47.90\%}& \textbf{51.23\%} & \textbf{ 74.24\%}\\ 
\midrule
VGG-Net \cite{Simonyan15}& 41.49\%& 43.46\%&  67.82\% \\ 
g-RACNN$_\text{VGG-Net}$ & 42.86\%& 44.72\%&  68.03\% \\ 
p-RACNN$_\text{VGG-Net}$  & \textbf{44.65\%}& \textbf{49.33\%} &  \textbf{ 69.17\%}\\ 
\midrule

 GoogLeNet \cite{szegedy2015going} & 46.85\% &  48.52\% &  69.28\% \\ 
 g-RACNN$_\text{GoogLeNet}$ & 50.37\% &  55.16\% & 69.77\% \\ 
 p-RACNN$_\text{GoogLeNet}$ &  \textbf{~50.76\%}& \textbf{ 57.30\%}&  \textbf{ 73.51\%}\\ 

\bottomrule
\end{tabular}}
\end{center}
\end{table}

\begin{figure*}[t]
\centering
\subfigure[AlexNet]{\includegraphics[width=0.32\linewidth, height=4.5cm]{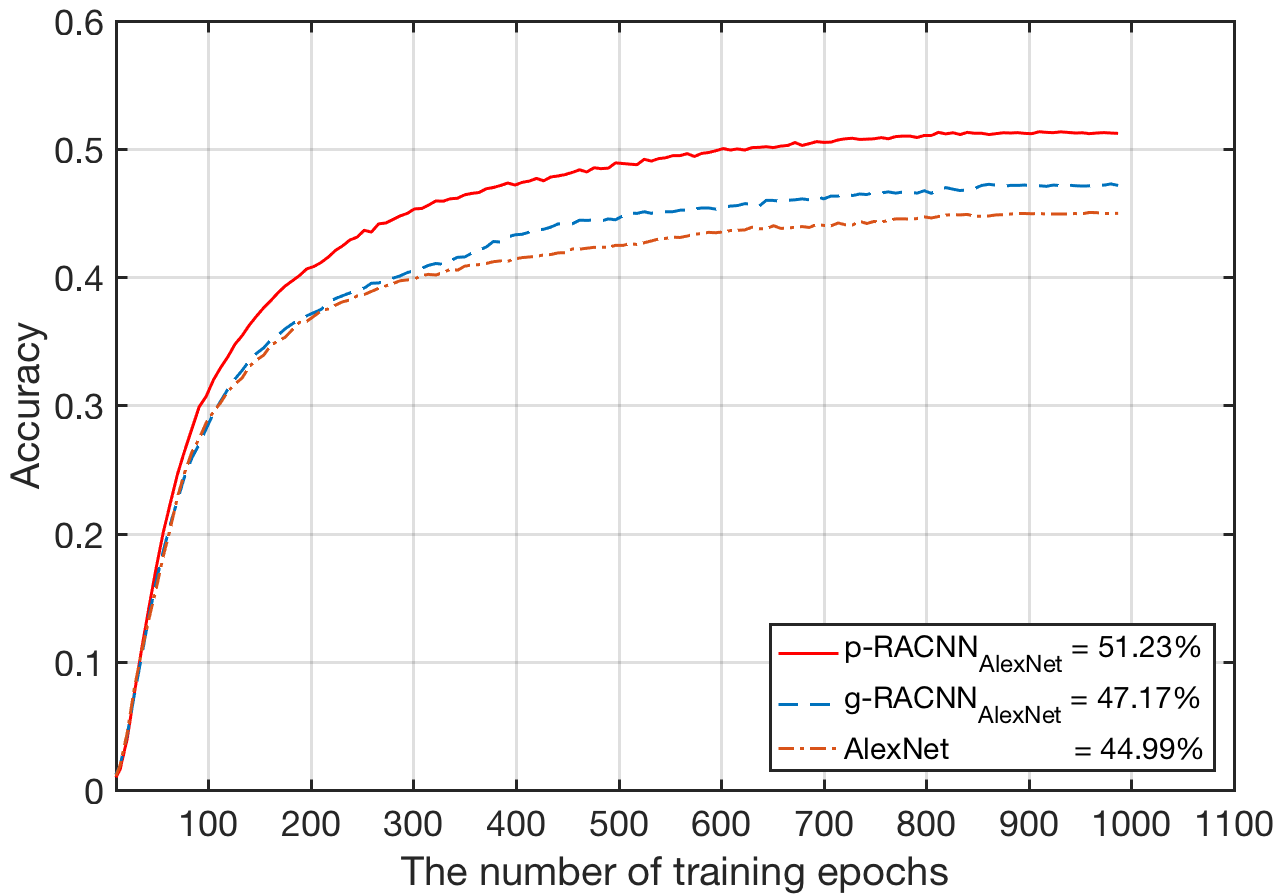}}
\subfigure[VGG-Net]{\includegraphics[width=0.32\linewidth, height=4.5cm]{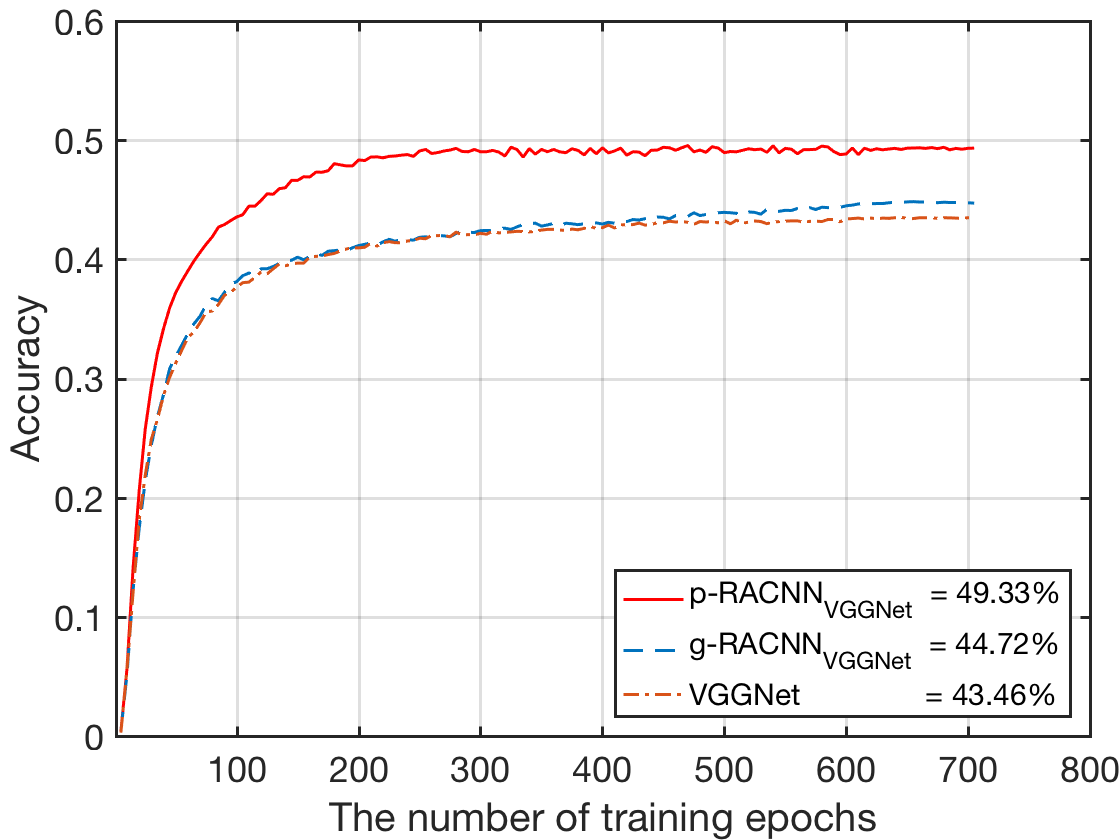}}
\subfigure[ GoogLeNet]{\includegraphics[width=0.32\linewidth, height=4.5cm]{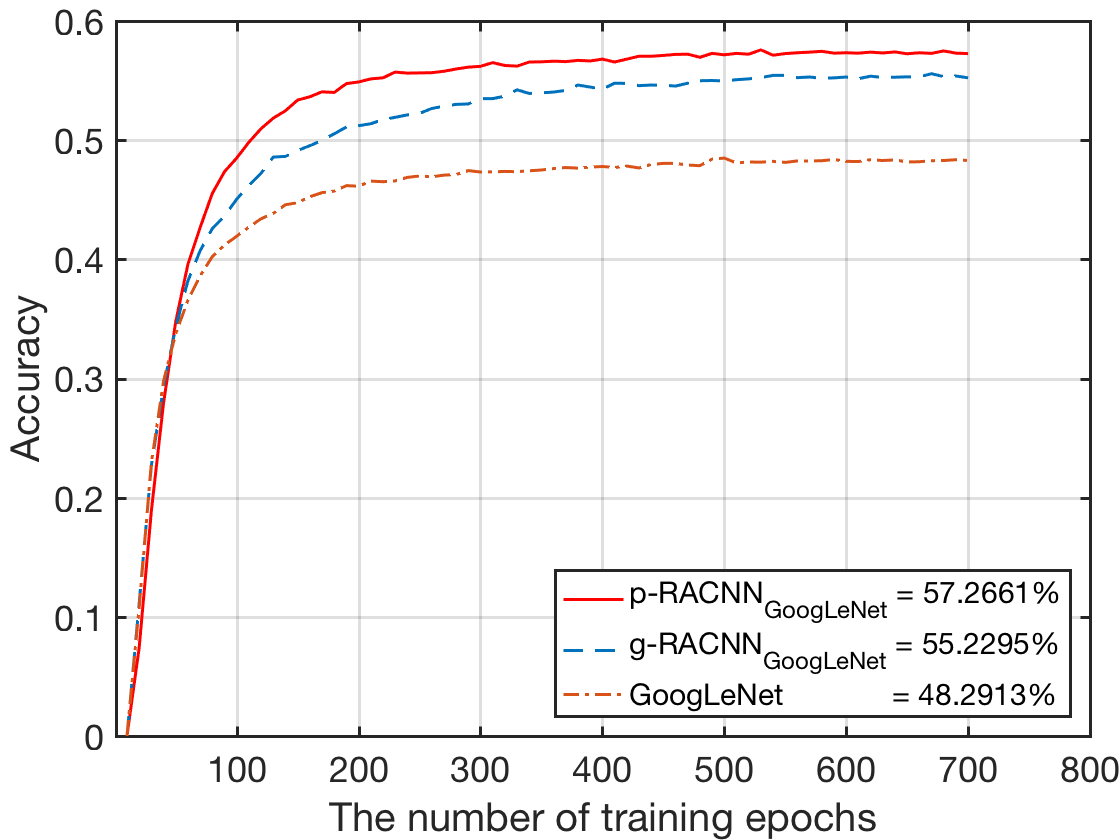}}
\caption{Training process of AlexNet, VGGNet  and GoogLeNet on the Caltech-UCSD Birds Dataset. }
\label{fig:birds-vgg}
\end{figure*}

In this experiment, we employ all layers in the AlexNet, VGG-Net  and GoogLeNet \cite{szegedy2015going} as
categorisation layers in RACNN. Note that, different from the
previous experiments, we freeze all
categorisation layers by setting learning rates and weights decays to
0 besides the last fully-connected layers of the baseline CNNs, and our
RACNN is then fine-tuned with low-resolution data.
Such setting treats categorisation layers in RACNN as an identical
classifier for evaluating the effect of adding convolutional SR
layers. RACNN with initial Gaussian and pre-trained weights are called
as g-RACNN and p-RACNN respectively. 
Comparative results are shown in Table~\ref{table:sr-layer} and
Fig.~\ref{fig:birds-vgg}.   
Both g-RACNN and p-RACNN consistently outperform the baseline CNNs in
all experiments.

With the same experimental setting except different initial weights
for convolutional super-resolution layers, the results of g-RACNN and
p-RACNN are reported. 
Test set accuracies in Table~\ref{table:sr-layer} and Fig.~\ref{fig:birds-vgg}
show that p-RACNN is superior to g-RACNN. 
p-RACNN and g-RACNN share the same network structure but differ only
in network weights initialisation of convolutional SR layers.  
In this sense, better performance of p-RACNN is credited to the
knowledge about refining low-resolution images (\ie pre-trained
weights), which verifies our motivation to boost low-resolution image
classification via image super-resolution. It is noteworthy that since
the feature extraction layers are frozen, the networks are not fine-tuned
to low-resolution specific features, but all performance boost are
owing to recovered high-resolution details important for classifcation
by the super-resolution layers.

\subsection{Evaluation on Varying Resolution }

\begin{table}[h]
\caption{Comparison with varying resolution level (Res. Level) on the Caltech-UCSD Birds 200-2011 Dataset. }
\label{table:cross-res}
\begin{center}
{\small \addtolength{\tabcolsep}{-1pt}
\begin{tabular}{lrrr}
\toprule
Res. Level & AlexNet \cite{krizhevsky2012imagenet}  & g-RACNN$_\text{AlexNet}$ &  p-RACNN$_\text{AlexNet}$   \\
\midrule
25$\times$25 & 31.58\% & 43.68\% & \textbf{45.06\%} \\ 
50$\times$50 & 44.99\% & 47.17\% & \textbf{51.23\%} \\
100$\times$100 & 51.01\% & 51.24\% & \textbf{52.88\%} \\
\bottomrule
\end{tabular}
}
\end{center}
\end{table}

We further evaluate our proposed RACNN method with respect to varying
resolutions on the Caltech-UCSD Birds 200-2011 Dataset. 
All low-resolution images are first up-scaled to the input image size, \ie 227$\times$227, before training models. 
The better performance of
RACNN$_\text{AlexNet}$ over conventional AlexNet is achieved for cross-resolution
fine-grained image classification, which is shown in Table
\ref{table:cross-res}.
   
We observe that our method performs much better for lower
resolution images (\eg 25$\times$25) than relatively high resolution
images (\eg 100$\times$100).  
In details,  p-RACNN$_\text{AlexNet}$ increases the accuracy by above
13\% for 25$\times$25 pixel images but less than 2\% improvement on
100$\times$100 resolution images.  
The reason is that the SR layers of RACNN play a significant role in
introducing texture details especially when missing more visual cues
of object classification in lower quality images, 
which further demonstrates our observation and motivation. 

In p-RACNN$_\text{AlexNet}$, the weights for convolutional SR layers are pre-trained only with 50$\times$50
resolution-level ImageNet images, but our RACNN is applied to varying resolution levels (\ie 25$\times$25 and 100$\times$100).
Further improvement on classification performance shows the generalisation of pre-trained weights for varying resolution levels, which demonstrates the generalisation ability of RACNN with pre-trained SR weights.


\section{Conclusion}\label{sec.conclusion}

We propose and verify a simple yet effective
resolution-aware convolutional neural network (RACNN) for fine-grained image
classification of low-resolution images.  
The results from extensive experiments indicate that the introduction of
convolutional super-resolution layers to conventional CNNs can indeed
recover fine details for low-resolution images and clearly boost
performance in low-resolution fine-grained classification. This result
can be explained by the fact that the super-resolution layers learn
to recover high resolution details that are important for classification
when trained end-to-end manner together with the classification layers.
The concept of our paper is generic and the existing convolutional
super-resolution and classification networks can be readily combined
to cope with low-resolution image classification.

%

\bibliographystyle{elsarticle-num}
\bibliography{RACNN}

\end{document}